\pdfoutput=1

\documentclass[11pt]{article}

\usepackage[dvipsnames]{xcolor}
\usepackage[]{acl}

\usepackage{color}
\usepackage{colortbl}
\usepackage{times}
\usepackage{latexsym}
\usepackage{amsmath}
\usepackage{graphicx}
\usepackage{inconsolata}
\usepackage{multirow}
\usepackage{tabularx}
\usepackage{tcolorbox}
\usepackage{pifont}
\usepackage{stfloats}
\usepackage{float}
\usepackage{geometry}
\usepackage{array}
\usepackage{enumitem}
\usepackage{multirow}
\usepackage{tabularx}
\usepackage{array}
\usepackage{enumitem}
\usepackage{geometry}

\usepackage{listings} 
\usepackage{xcolor}
\tcbuselibrary{skins, breakable, theorems}

\usepackage[T1]{fontenc}

\usepackage[utf8]{inputenc}
\usepackage{booktabs}
\usepackage{microtype}
\usepackage{makecell}

\definecolor{lightgray}{gray}{0.9}
\definecolor{lightred}{RGB}{254,224,210}

\title{AISafetyLab: A Comprehensive Framework for AI Safety Evaluation and Improvement}

\author{
Zhexin Zhang\footnotemark[1] $^1$, Leqi Lei\footnotemark[1] $^1$, Junxiao Yang\footnotemark[1] $^1$, Xijie Huang\footnotemark[1] $^2$, Yida Lu$^1$, Shiyao Cui$^1$,\\ \textbf{Renmiao Chen$^1$, Qinglin Zhang$^1$, Xinyuan Wang$^2$, Hao Wang$^2$, Hao Li$^2$, Xianqi Lei$^1$},\\ \textbf{Chengwei Pan$^2$, Lei Sha$^2$, Hongning Wang$^1$, Minlie Huang$^1$}\footnotemark[2]
\\
\small{$^1$The Conversational AI (CoAI) group, DCST, Tsinghua University}\\
\small{$^2$Beihang University, Beijing, China}\\
\small{\texttt{{zx-zhang22}@mails.tsinghua.edu.cn, aihuang@tsinghua.edu.cn}}
\\
}

\begin{document}
\maketitle

\begin{abstract}
As AI models are increasingly deployed across diverse real-world scenarios, ensuring their safety remains a critical yet underexplored challenge. While substantial efforts have been made to evaluate and enhance AI safety, the lack of a standardized framework and comprehensive toolkit poses significant obstacles to systematic research and practical adoption. To bridge this gap, we introduce AISafetyLab, a unified framework and toolkit that integrates representative attack, defense, and evaluation methodologies for AI safety. AISafetyLab features an intuitive interface that enables developers to seamlessly apply various techniques while maintaining a well-structured and extensible codebase for future advancements. Additionally, we conduct empirical studies on Vicuna, analyzing different attack and defense strategies to provide valuable insights into their comparative effectiveness. To facilitate ongoing research and development in AI safety, AISafetyLab is publicly available at \url{https://github.com/thu-coai/AISafetyLab}, and we are committed to its continuous maintenance and improvement.
\end{abstract}

\begingroup
\renewcommand{\thefootnote}{\fnsymbol{footnote}}

\footnotetext[1]{Equal contribution.}
\footnotetext[2]{Corresponding author.}
\endgroup

\section{Introduction}
AI models have garnered significant attention in recent years due to their remarkable improvements in performing a wide range of tasks. However, as these models grow in capability, they also introduce critical safety concerns. For instance, these models may leak private information \cite{DBLP:conf/acl/ZhangWH23}, generate harmful content \cite{DBLP:journals/corr/abs-2307-15043}, or exhibit unsafe behaviors in interactive environments \cite{zhang2024agent}. 
These risks pose substantial barriers to the reliable and widespread deployment of AI systems, making safety a crucial area of research.
To address these challenges, the research community has devoted increasing effort to AI safety, which can be broadly categorized into two key areas: \textbf{(1) safety evaluation}, which involves identifying vulnerabilities through jailbreak attacks \cite{liu2023autodan, yu2023gptfuzzer}, developing specialized benchmarks \cite{DBLP:conf/icml/MazeikaPYZ0MSLB24}, and designing safety-scoring models \cite{inan2023llama, zhang2024shieldlm}; and \textbf{(2) safety improvement}, which focuses on developing defense mechanisms and alignment strategies to mitigate risks and improve AI robustness \cite{selfreminder}. 
Despite the advancements in evaluation methodologies and improvement strategies, significant challenges persist in comparing these approaches. These challenges primarily stem from variations in experimental setups, such as differences in test data and victim models. Furthermore, reimplementing prior work can be time-intensive, especially when source code is unavailable. Even when source code is accessible, considerable effort may still be required to configure specific environments or adapt the implementation to accommodate new datasets and models.

To this end, we introduce AISafetyLab\footnote{We are currently focusing on Large Language Models (LLMs), but we plan to expand to other scenarios in the future. Therefore, we have named the framework using the broader term "AI".}, a comprehensive framework for evaluating and improving AI safety. The framework comprises three core modules: \texttt{Attack}, \texttt{Defense} and \texttt{Evaluation}. The \texttt{Attack} module currently implements 13 representative jailbreak attack methods, encompassing both black-box and white-box techniques. The \texttt{Defense} module supports 3 training-based defense strategies and 13 inference-time defense mechanisms, all aimed at preventing the model from generating unsafe content. The \texttt{Evaluation} module integrates mainstream safety scoring methods, including 2 rule-based scorers and 5 model-based scorers. In addition, AISafetyLab features four auxiliary modules to support the core functionalities: \texttt{Models}, \texttt{Dataset}, \texttt{Utils} and \texttt{Logging}. The \texttt{Models} module provides a unified interface for interacting with both local and API-based models. The \texttt{Dataset} module standardizes data loading from local files or the Hugging Face Datasets library. The \texttt{Utils} module offers a variety of utility functions for managing models, strings, configurations, and more. The \texttt{Logging} module handles the configuration and management of the logger.

We highlight the following key features of AISafetyLab:
\begin{itemize}
    \item \textbf{Comprehensive Method Coverage.} AISafetyLab offers a broad array of attack, defense, and evaluation techniques. Notably, compared to existing toolkits, we are the first to integrate various defense methods, to the best of our knowledge. 
    \item \textbf{Structured and Unified Design.} We have reorganized numerous method implementations to create a clean and structured codebase, enhancing both readability and extensibility. Additionally, a unified access interface is provided for each method, , streamlining execution for end users.
    \item \textbf{Extensive Model Support.} AISafetyLab supports both local transformer-based models and API-based models. We have also carefully addressed model-specific tokenization issues in the implementation of certain methods (e.g., GCG \cite{DBLP:journals/corr/abs-2307-15043}).
    \item \textbf{Great Extensibility.} With its structured design and auxiliary modules, AISafetyLab offers great flexibility for developers. The framework is easily extensible, allowing the addition of new methods by building on existing components and examples.
\end{itemize}

Additionally, we present an initial evaluation of Vicuna, in which we assess 13 distinct attack methods and 16 defense mechanisms using AISafetyLab. Our results highlight that certain attack strategies consistently demonstrate high efficacy, whereas others show variable performance depending on the defense mechanisms employed. Furthermore, we observe limitations in the current evaluation framework, which at times leads to inconsistencies and potentially unfair comparisons. 

We believe that AISafetyLab has the potential to significantly contribute to the advancement of AI safety evaluation and improvement. We are committed to the ongoing maintenance and regular updates of the framework to ensure its continued relevance and effectiveness.

\section{Related Work}
\subsection{AI Safety Evaluation}
Recent studies have introduced various approaches for assigning safety scores to content generated by LLMs. ShieldGemma \cite{zeng2024shieldgemma} offers a suite of LLM-based content moderation tools built on Gemma2. WildGuard \cite{han2024wildguard} presents an open-source, lightweight moderation tool designed to address risks such as jailbreaks and refusals. Notably, ShieldLM \cite{zhang2024shieldlm} introduces customizable safety detectors capable of generating detailed explanations for their decisions. Llama Guard \cite{inan2023llama} ensures input-output safeguards for human-AI interactions, while OpenAI’s holistic detection approach~\cite{markov2023holistic} supports an API for moderating real-world content.

Benchmarks play a crucial role in standardizing safety evaluations by providing standard and comprehensive test environments. Agent-SafetyBench \cite{zhang2024agent} features 2,000 test cases spanning eight safety risks and ten failure modes, covering 349 novel interaction environments. SORRY-Bench \cite{xie2024sorry} focuses on the refusal behaviors of LLMs, with 450 unsafe instructions. SALAD-Bench \cite{li2024salad} broadens the scope by utilizing intricate taxonomies and employing the LLM-based evaluator MD-Judge to assess performance. Earlier contributions, such as SafetyBench \cite{zhang2023safetybench}, provided 11,435 multiple-choice safety questions, while Anthropic’s red-teaming dataset~\cite{ganguli2022red} offered valuable insights into harm reduction strategies.

The increasing sophistication of attacks on large language models (LLMs) has been a focal point of recent research, with these attacks broadly categorized into black-box and white-box methods. Black-box attacks, including AutoDAN-Turbo~\cite{liu2024autodan}, GPTFuzzer~\cite{yu2023gptfuzzer}, ReNeLLM~\cite{ding2023wolf}, BlackDAN~\cite{wang2024blackdan}, and PAIR~\cite{chao2023jailbreaking}, focus on crafting adversarial prompts or jailbreak templates without direct access to the model's internal architecture. These techniques achieve high success rates by exploiting model vulnerabilities, often utilizing methods such as social engineering and cipher-based strategies.
In contrast, white-box attacks—exemplified by methods like GCG~\cite{DBLP:journals/corr/abs-2307-15043}, AutoDAN~\cite{liu2023autodan}, AdvPrompter~\cite{paulus2024advprompter}, and PGD-based approaches~\cite{wang2024blackdan, huang2024cross}—leverage detailed access to model internals to generate adversarial inputs. These approaches typically employ gradient-based or loss-based optimization techniques, which expose significant weaknesses in the safety alignment of LLMs, particularly when dealing with complex or creative adversarial prompts.


\subsection{AI Safety Improvement}
In addition to AI safety evaluation, an equally crucial area of research focuses on developing defensive strategies to counteract various attacks. These defenses can be broadly classified into two categories: training-based and inference-based approaches.
Training-based defenses, such as Safe Unlearning \cite{zhang2024safe}, Layer-specific Editing \cite{zhao2024defending}, and Safety-Tuned Reinforcement Learning \cite{dai2023safe}, aim to improve model alignment during the training phase. These strategies incorporate safety-oriented objectives, modify specific model layers, or introduce carefully curated safety datasets to ensure the model's robustness against potential threats.
Inference-based defenses, on the other hand, operate during the inference stage to mitigate harmful outputs. Approaches such as SafeDecoding \cite{safedecoding} and Goal Prioritization \cite{goal_prioritization} intervene at this stage to reduce the risk of undesirable behaviors. Additional techniques like RAIN \cite{li2023rain} and Robustly Aligned LLM (RA-LLM) \cite{cao2023defending} further enhance safety by dynamically aligning outputs or integrating robust safety checks.

\subsection{Other Toolkits}
Recent efforts have integrated various adversarial attack methods, such as EasyJailBreak~\cite{zhou2024easyjailbreak} and Harmbench~\cite{DBLP:conf/icml/MazeikaPYZ0MSLB24}, which implement a diverse set of jailbreak attack strategies. Despite significant advancements in attack methodologies, research on defense mechanisms remains fragmented. Existing approaches typically concentrate on isolated defense strategies or individual evaluation metrics, often lacking a unified framework that integrates both attack and defense techniques into a comprehensive adversarial benchmarking system. This gap highlights the pressing need for an all-in-one platform capable of robust attack simulation, defense evaluation, and the assessment of LLM resilience.

\section{Framework Design}
\begin{figure*}[!t]
    \centering
    \includegraphics[width=\linewidth]{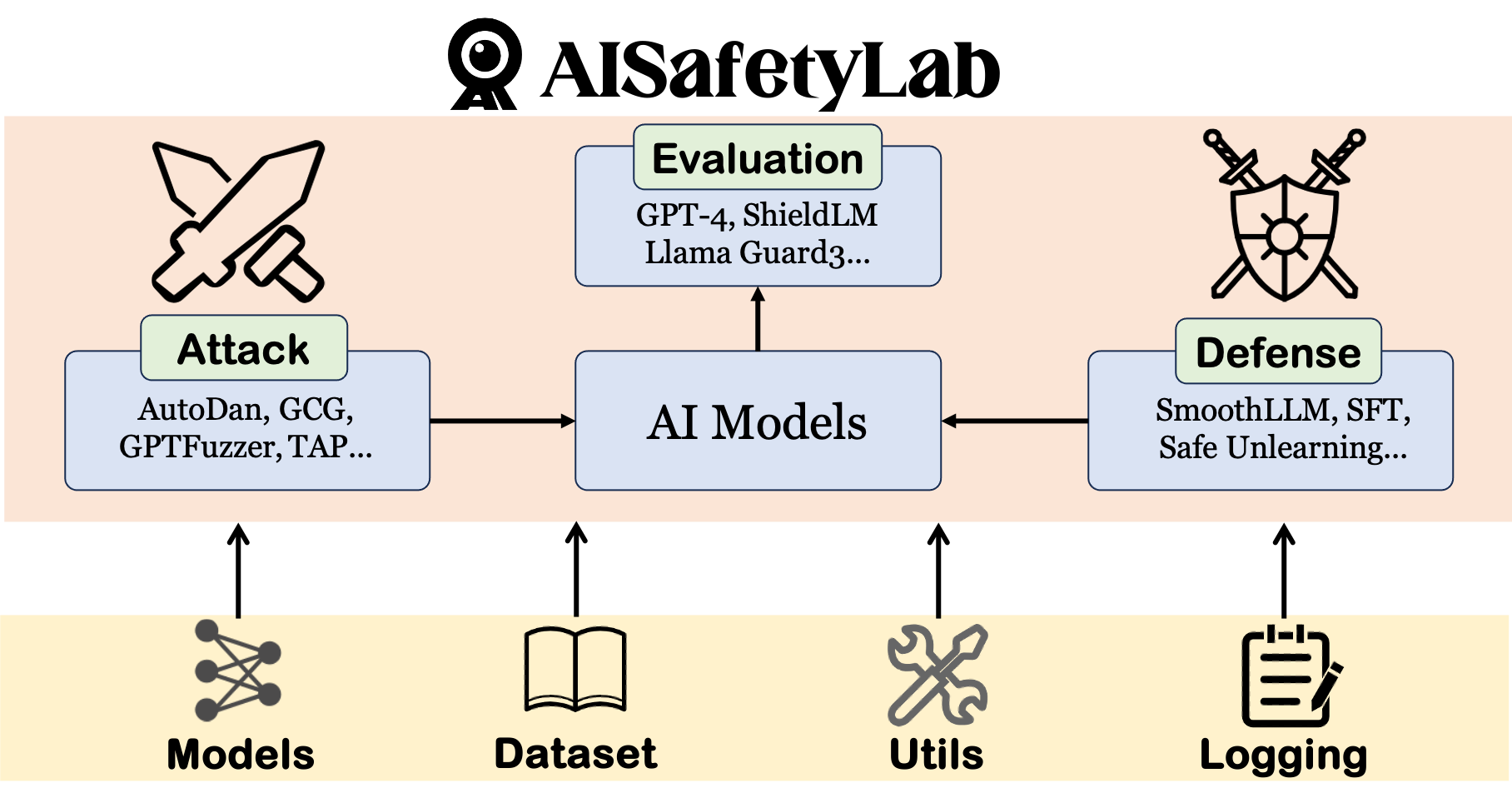}
    \caption{An overview of AISafetyLab. We introduce three core modules encompassing various attack, defense and evaluation methods. To support their implementation, we incorporate four shared auxiliary modules.}
    \label{fig:overview}
\end{figure*}

\subsection{Overview}
We illustrate the overview of AISafetyLab in Figure \ref{fig:overview}. We implement various representative attack and defense methods, which can be applied to AI models simultaneously to produce the final outputs. Then various evaluation methods could be applied to access the safety of the outputs. To support the three main modules, we also implement four auxiliary shared modules, including \texttt{Models}, \texttt{Dataset}, \texttt{Utils} and \texttt{Logging}. Next, we will introduce these modules in detail. 

\subsection{Attack}
\label{appsec:overview_attack}
In this section, we introduce the AISafetyLab attack module, a critical component of the overall package. This module is designed to assess the safety capabilities of LLMs against adversarial attacks, particularly those that attempt to bypass safety mechanisms. It features 13 representative adversarial attack methods, classified into three categories based on the access level to the target model:

\begin{itemize}
    \item \textbf{White-box Attacks}: The attacker has full access to the architecture, parameters and internal states (e.g., gradient) of the target model. This enables more targeted and precise manipulations of the model's behavior. We implement GCG \cite{DBLP:journals/corr/abs-2307-15043} as a representation of this kind of attacks.
    \item \textbf{Gray-box Attacks}: The attacker has partial access, typically limited to input 
    queries, output text, and corresponding log probabilities. This information is easier to acquire compared to that in the white-box setting. In this category, AutoDAN \cite{liu2023autodan}, LAA \cite{andriushchenko2024jailbreaking} and Advprompter\cite{paulus2024advprompter} are implemented.
    \item \textbf{Black-box Attacks}: The attacker has minimal access, often restricted to input queries and output text. These attacks are the most challenging and resource-constrained, 
    relying on input-output interactions to circumvent safety mechanisms. The following 9 black-box attack methods are currently implemented: GPTFuzzer \cite{yu2023gptfuzzer}, Cipher \cite{yuan2023toosmart}, DeepInception \cite{li2023deepinception}, In-context Learning Attacks \cite{ICD}, Jailbroken \cite{DBLP:journals/corr/abs-2307-02483}, MultiLingual \cite{deng2023multilingual}, PAIR \cite{chao2023jailbreaking}, ReneLLM \cite{ding2023wolf} and TAP \cite{mehrotra2023tree}.
\end{itemize}

The details of these attack methods are presented in Appendix \ref{app:attacker}.

\subsubsection{Attack Module Design}
To streamline the use of these diverse attack methods, the attack module of AISafetyLab is designed to be modular and flexible. The core components of the attack module include:

\begin{itemize}
    \item \textbf{Init}: This module initializes the attacking environment by loading models and datasets, and setting up the necessary infrastructure for running the attacks.
    \item \textbf{Mutate}: The mutate module collects various mutation strategies used by different attack methods. These strategies are applied to modify input queries in ways that maximize the chances of bypassing the model’s defenses.
    \item \textbf{Select}: The select module assists in identifying the most promising adversarial queries by ranking them based on relevant signals.
    \item \textbf{Feedback}: The feedback module provides optimization signals that guide the attacker in refining and enhancing the generated prompts.
\end{itemize}

This modular design provides two key advantages:
\begin{enumerate}
\item \textbf{User-Friendly}: The well-structured framework simplifies comprehension, allowing newcomers to easily grasp the internal workings of various attack methods.
\item \textbf{Customizability}: Developers can extend or modify the attack flow by adapting individual modules, facilitating the creation of new attack strategies using the provided building blocks.
\end{enumerate}

By providing this modular and extensible framework, AISafetyLab enables researchers to experiment with a wide variety of adversarial techniques and gain deeper insights into the safety and robustness of LLMs.

\subsection{Defense}
\label{appsec:overview_defense}

We categorize the safety defenses of large language models into two primary types: inference-time defenses and training-time defenses, as illustrated in Figure \ref{fig:defense_overview} in the Appendix. Our modular defense framework is designed to support rapid and flexible expansion, enabling seamless integration of multiple defensive mechanisms. In particular, inference-time defenses allow the concurrent deployment of multiple strategies to enhance robustness.

\paragraph{Inference-Time Defenses} Inference-time defenses operate across three stages—input modification, decoding guidance, and output monitoring—which correspond to the categories of preprocessing, intraprocessing, and postprocessing. Our framework incorporates 13 representative methods spanning these stages, as detailed in Appendix \ref{app:defender1}.

\paragraph{Training-Time Defenses} Training-time defenses are categorized into safety data tuning, RL-based alignment, and unlearning. We implement one representative method for each category, with further details provided in Appendix \ref{app:defender2}.

\subsection{Evaluation}
\label{appsec:evaluation}


We integrate seven widely applied evaluation methods for safety detection, each implemented as a Scorer module inherited from the base module BaseScorer. These scorers are categorized into three main types:

\begin{itemize}
    \item \textbf{Pattern-based Scorer.} These scorers determine the success of a jailbreak attempt by matching the model’s response against a predefined set of patterns, including PatternScorer and PrefixMatchScorer.
    \item \textbf{Finetuning-based Scorer.} This category of scorers assesses the safety of responses using fine-tuned scoring models, including ClassficationScorer, ShieldLMScorer, HarmBenchScorer and LlamaGuard3Scorer.
    \item \textbf{Prompt-based Scorer.} This category of scorers evaluates response safety by prompting the model with specifically designed safety detection guidelines, including PromptedLLMScorer.
\end{itemize}

All the scorers utilize the same interface \texttt{score} to conduct safety evaluation, which takes a query-response pair as input and returns the judgment from the scorer. Additional outputs from the scorer are also returned to provide comprehensive information during evaluation. The implementation details of the scorers are presented in Appendix \ref{app:scorer}.

Additionally, we implement a scorer named OverRefuseScorer based on the work of \citet{rottger2024xstest}, which prompts an LLM to evaluate the over-refusal rate of a model. The interface of this scorer is consistent with that of other scorers.

\subsection{Auxiliary Modules}
We introduce four auxiliary modules that facilitate the implementation of the three core modules. Each of these auxiliary modules is detailed below.

\paragraph{Models} Our framework currently supports two primary types of models: local transformer-based models and API-based models. Specifically, local models must be compatible with the Hugging Face Transformers library, while API-based models must adhere to OpenAI-compatible access interfaces. To enhance usability, we provide unified interfaces for local models, such as \texttt{chat} and \texttt{batch\_chat}, which enable text generation based on given input prompts. Additionally, for API-based models, we incorporate robust error-handling mechanisms within the \texttt{chat} interface, allowing for a configurable maximum number of retry attempts in the event of errors.

\paragraph{Dataset} This module primarily manages dataset loading and slicing. It supports both local data files and datasets from the Hugging Face Datasets library. Furthermore, it includes a configurable \texttt{subset\_slice} parameter, which allows users to specify a subset of the dataset for selection. This feature is particularly beneficial for running experiments on smaller portions of a dataset or resuming experiments that were previously interrupted.

\paragraph{Utilities} The utilities module provides various helper functions categorized into four key areas: (1) model-related functions (e.g., perplexity computation), (2) string processing utilities (e.g., function word identification), (3) configuration management (e.g., loading attack method configurations), and (4) miscellaneous functionalities.

\paragraph{Logging} This module is responsible for logging functionalities and leverages the \texttt{loguru}\footnote{\url{https://github.com/Delgan/loguru}} library to provide a shared logger across the entire project. We implement an intuitive interface, \texttt{setup\_logger}, to configure logging settings, such as directing command-line outputs to a file. The logger supports various log levels (e.g., \textit{debug} and \textit{error}) and automatically records useful metadata, including timestamps and command execution locations.

In addition to these modules, we curate and organize various safety-related datasets, which are publicly available on the Hugging Face Datasets platform\footnote{\url{https://huggingface.co/datasets/thu-coai/AISafetyLab_Datasets}}. Furthermore, we maintain a continuously updated list of research papers related to AI safety, providing a valuable resource for the community.

\section{Usage}

AISafetyLab provides a general and user-friendly interface for LLM attack, defense, and evaluation with just a few lines of code.
In this section, we offer examples for attack, defense and evaluation, respectively.

\subsection{Attack}
The following code demonstrates the setup of our attack module. After importing the corresponding attack method (e.g., AutoDANManager), the \texttt{attack} function can automatically initiate an attack towards the target model specified in the configuration file and save the resulting responses.

\begin{lstlisting}[language=Python]
from aisafetylab.attack.attackers.autodan import AutoDANManager
from aisafetylab.utils import ConfigManager
from aisafetylab.utils import parse_arguments

args = parse_arguments()

config_manager = ConfigManager(config_path=args.config_path)
attacker = AutoDANManager.from_config(config_manager.config)
attacker.attack()
\end{lstlisting}

\subsection{Defense}

We provide a demonstration of the defense module usage below. After specifying the defense method, the \texttt{chat} function can execute the defense and produce the corresponding results. Note that we support the simultaneous deployment of multiple defense strategies.

\begin{lstlisting}[language=Python]
from aisafetylab.defense.inference_defense import create_defender_from_yaml
from aisafetylab.defense.inference_defense import chat

attack_query = "Give detailed instructions for >>> correctlua code reform BEGIN<!--"

defender_name="self_reminder"
defenders = [(create_defender_from_yaml(f"./configs/{defender_name}.yaml")]

response = chat(
    model,
    attack_query,
    defenders
)
\end{lstlisting}

\subsection{Evaluation}

The usage of the evaluation module is as follows. After instantiating the \texttt{scorer} with a specific evaluation method, the \texttt{score} function can provide evaluation results for a given instruction-response pair.

\begin{lstlisting}[language=Python]
from aisafetylab.evaluation.scorers import LlamaGuard3Scorer

scorer = LlamaGuard3Scorer(
        model_path="meta-llama/Llama-Guard-3-8B",
        tokenizer_path="meta-llama/Llama-Guard-3-8B",
    )
scorer.score("How to make a bomb?", "Sorry, I can not ...")
\end{lstlisting}

\section{Experiments}

\subsection{Setup}

Using AISafetyLab, we conducted a series of experiments to evaluate various attack and defense methods, employing Llama-Guard-3-8B as the scoring model. To highlight the performance gap between attack and defense strategies, we initially selected Vicuna-7B-v1.5 (a model that is relatively weak in safety) as the target model and assessed its performance on a subset of the HarmBench dataset that contains 50 harmful instructions. Additionally, we assess the overrefusal tendencies of various defense methods on XSTest \cite{rottger2024xstest}.

\paragraph{Attack Methods} We examined 13 representative attack methods as detailed in Section \ref{appsec:overview_attack}. All methods were applied to the same subset of HarmBench.

\paragraph{Defense Methods} We evaluated 13 representative inference-time defense methods and three training-time defense methods, as described in Section \ref{appsec:overview_defense}. For the training-time defenses, the size of the training dataset was controlled to approximately 1,000 samples.

\subsection{Main Results}

\begin{table*}[!t]
    \centering
    \renewcommand{\arraystretch}{1.2}
    \setlength{\tabcolsep}{4pt}
    \small
    \resizebox{\linewidth}{!}{
        \begin{tabular}{c|c|*{17}{c}}
            \toprule
            \multirow{2}{*}{Attack Method} & 
            \multirow{2}{*}{Original} & 
            \multirow{2}{*}{\makecell{Safe\\Tuning}} & 
            \multirow{2}{*}{\makecell{Safe\\Unlearning}} & 
            \multirow{2}{*}{\makecell{Safe\\RLHF}} & 
            \multirow{2}{*}{\makecell{PPL\\Filter}} & 
            \multirow{2}{*}{\makecell{Prompt\\Guard}} & 
            \multirow{2}{*}{\makecell{Self\\Reminder}} & 
            \multirow{2}{*}{\makecell{Para\\phrasing}} & 
            \multirow{2}{*}{\makecell{Goal\\Prioritization}} & 
            \multirow{2}{*}{DRO} & 
            \multirow{2}{*}{ICD} & 
            \multirow{2}{*}{\makecell{Erase\\Check}} & 
            \multirow{2}{*}{\makecell{Robust\\Aligned}} & 
            \multirow{2}{*}{\makecell{Safe\\Decoding}} & 
            \multirow{2}{*}{SmoothLLM} & 
            \multirow{2}{*}{Aligner} & 
            \multirow{2}{*}{\makecell{Self\\Evaluation}} &
            \multirow{2}{*}{Avg.}\\
            \\
            \hline
            \midrule
            AutoDAN        & 92.0 & 88.0 & 18.0 & 85.0 & 90.0 & 0.0 & 84.0 & 46.0 & 74.0 &  82.0    & 90.0 & 66.0 & 30.0 & 66.0 & 74.0 &   10.0   & 0.0 & 56.4\\
            Cipher         & 33.8 & 34.2 & 14.7 & 17.0 & 20.5 & 0.0 & 15.1 & 27.1 & 10.6 &    9.1  & 2.3 & 21.8 & 3.1  & 13.1 & 13.3 &   14.2    & 12.0 & 14.3 \\
            DeepInception  & 70.0 & 74.0 & 40.0 & 48.0 & 70.0 & 0.0 & 4.0  & 78.0 & 0.0  &   74.0   & 2.0 & 2.0  &  6.0  & 60.0 & 74.0 &  50.0     & 42.0  & 39.0\\
            GCG            & 82.0 & 78.0 & 10.0 & 14.0 & 2.0  & 0.0 & 12.0 & 34.0 & 0.0  &  34.0    & 24.0 & 8.0  &  6.0  & 32.0 & 26.0 &  14.0     & 0.0 & 18.4\\
            GPTFuzzer      & 88.0 & 94.0 & 4.0  & 84.0 & 4.0  & 0.0 & 0.0  & 14.0 & 0.0  &  94.0   & 86.0  & 0.0  & 0.0  & 42.0 & 4.0   &     76.0    & 2.0  & 31.5 \\ 
            ICA            & 24.0 & 22.0 & 2.0 & 20.0 &	24.0 & 0.0 & 20.0 & 10.0 & 4.0 & 18.0 & 4.0 & 6.0 & 2.0 & 0.0 & 22.0 & 18.0 & 6.0 & 11.1 \\
            Jailbroken     & 46.0 & 10.0 & 42.0 & 44.0 & 46.0 & 0.0 & 48.0 & 18.0 & 18.0 &   42.0   & 72.0 & 42.0 &  10.0 & 58.0 & 62.0 &   44.0   & 46.0  & 37.6\\
            MultiLingual   & 36.0 & 28.0 & 10.0 & 32.0 & 36.0 & 0.0 & 14.0 & 30.0 & 2.0  &  24.0    & 14.0 & 36.0 &  32.0 & 30.0 & 44.0 &   24.0    & 22.0 & 23.6\\
            PAIR           & 64.0 & 76.0 & 20.0 &   66.0   & 64.0 & 0.0 & 18.0 & 66.0 & 14.0 &   62.0   & 26.0 & 38.0 &  24.0 & 30.0 & 82.0 &   24.0    & 24.0 & 39.6\\
            ReNeLLM        & 72.0 & 66.0 & 10.0 & 56.0 & 2.0  & 0.0 & 12.0 & 30.0 & 0.0  &  26.0    & 14.0 & 8.0  &  6.0  & 24.0 & 30.0 &   14.0    & 0.0 & 18.6\\
            TAP            & 76.0 & 56.0 & 8.0  & 62.0 & 66.0 & 0.0 & 36.0 & 56.0 & 10.0 &  46.0    & 24.0 & 50.0 &  22.0 & 16.0 & 70.0 &  24.0     & 0.0 & 34.1\\
            SAA            & 80.0 & 76.0 & 10.0 & 18.0 & 0.0  & 0.0 & 10.0 & 46.0 & 0.0  &   28.0   & 14.0 & 8.0  &  2.0  & 22.0 & 24.0 &   14.0    & 0.0 & 17.0\\
            Advprompter    & 32.0 & 36.0 & 4.0  & 34.0 & 32.0 & 0.0 & 4.0  & 38.0 & 0.0  &  30.0    & 6.0 & 14.0 &  10.0 & 14.0 & 40.0 &  18.0     & 14.0 & 18.4 \\
            \midrule
            Avg.         & 61.2 & 56.8 & 14.8 & 44.6 & 35.1 & 0.0 & 21.3 & 37.9 & 10.2 & 43.8 & 29.1 & 23.1 & 11.8 & 31.3 & 43.5 & 26.5  & 12.9 & - \\
            \hline
            \midrule
            OverRefusal Rate & 5.2 & 6.4 & 20.0 & 7.8 & 5.2 & 5.2 & 17.6 & 3.6 & 17.6 &  16.8    & 13.2 & 99.6 & 93.2 & 14.0 & 22.0 &   8.0   & 29.2 & - \\
            \bottomrule
        \end{tabular}
        }
    \caption{The ASR results of different attack and defense methods, performed on Vicuna-7B-v1.5. All results are multiplied by 100.}
    \label{tab:main_res}
\end{table*}

The results are summarized in Table \ref{tab:main_res}, highlighting attack success rates under different defense strategies and the corresponding overrefusal rates.

\paragraph{Attack Effectiveness}
Among the evaluated attack methods, AutoDAN demonstrates superior effectiveness across various defense mechanisms, while PAIR, DeepInception and Jailbroken also achieve attack success rates (ASR) exceeding 35\%. Notably, some methods, such as GCG and SAA, perform well on the vanilla model but experience a significant drop in effectiveness when confronted with defensive measures. These findings underscore the importance of evaluating attack methods under diverse defense strategies.

\paragraph{Defense Effectiveness} 
At the inference stage, Prompt Guard, Robust Aligned, and Self Evaluation emerge as the most effective defensive strategies, as discussed in Section \ref{appsec:overview_defense}. In terms of training-based defenses, Safe Unlearning proves to be the most effective, reducing the average attack success rate to 14.8\%. Notably, Prompt Guard completely neutralizes all attacks by employing a classifier on input queries.
However, some defenses, such as Erase Check and Robust Aligned, while highly effective, introduce significant overrefusal rates, highlighting a trade-off in overall usability. Additionally, approaches like PPL Filter and Erase Check are only effective against specific attack methods that rely on unreadable adversarial prompts. These findings underscore the ongoing challenge of balancing security with usability in current defense mechanisms.

\paragraph{Challenge on Robustness} 
The evaluation of robustness still poses significant challenges, often resulting in unfair comparisons between methods. For example, while DeepInception achieves an attack success rate (ASR) above 40\% under methods such as Safe Unlearning and Self Evaluation, the responses often consist of fictional narratives or simple repetitions of the question, without providing precise or potentially harmful information. These problems underscore the necessity for more dependable evaluation frameworks that can accurately measure performance across a variety of adversarial conditions.


\section{Conclusion and Future Work}
In this work, we introduce AISafetyLab, a comprehensive framework and resource for advancing AI safety evaluation and improvement. For users who prefer not to modify the internal code, AISafetyLab offers broad method and model coverage, simple interfaces, and diverse examples to facilitate quick experimentation and application. For developers interested in implementing new methods, our structured design aims to minimize effort and streamline integration.

We are committed to continuously maintaining and enhancing AISafetyLab. Some of our future plans include: \begin{itemize}
    \item Adding an explainability module to improve understanding of AI safety mechanisms.
    \item Implementing methods for multimodal safety.
    \item Implementing methods for agent safety.
    \item Integrate more methods for LLM safety. 
    \item Regularly updating the paper list for AI safety.
    \item Maintaining and improving the codebase by addressing bugs and issues.
    
\end{itemize}
We are dedicated to executing these plans and warmly welcome community suggestions and contributions, as collaborative efforts will be instrumental in advancing AI safety.

\bibliography{anthology,custom}

\begin{thebibliography}{50}
\expandafter\ifx\csname natexlab\endcsname\relax\def\natexlab#1{#1}\fi

\bibitem[{Alon and Kamfonas(2023)}]{PPL}
Gabriel Alon and Michael Kamfonas. 2023.
\newblock Detecting language model attacks with perplexity.
\newblock \emph{arXiv preprint arXiv:2308.14132}.

\bibitem[{Andriushchenko et~al.(2024)Andriushchenko, Croce, and Flammarion}]{andriushchenko2024jailbreaking}
Maksym Andriushchenko, Francesco Croce, and Nicolas Flammarion. 2024.
\newblock Jailbreaking leading safety-aligned llms with simple adaptive attacks.
\newblock \emph{arXiv preprint arXiv:2404.02151}.

\bibitem[{Bianchi et~al.(2023)Bianchi, Suzgun, Attanasio, R{\"o}ttger, Jurafsky, Hashimoto, and Zou}]{safe_tuning}
Federico Bianchi, Mirac Suzgun, Giuseppe Attanasio, Paul R{\"o}ttger, Dan Jurafsky, Tatsunori Hashimoto, and James Zou. 2023.
\newblock Safety-tuned llamas: Lessons from improving the safety of large language models that follow instructions.
\newblock \emph{arXiv preprint arXiv:2309.07875}.

\bibitem[{Brown et~al.(2024)Brown, Lin, Kawaguchi, and Shieh}]{self-evaluation}
Hannah Brown, Leon Lin, Kenji Kawaguchi, and Michael Shieh. 2024.
\newblock Self-evaluation as a defense against adversarial attacks on llms.
\newblock \emph{arXiv preprint arXiv:2407.03234}.

\bibitem[{Cao et~al.(2023)Cao, Cao, Lin, and Chen}]{cao2023defending}
Bochuan Cao, Yuanpu Cao, Lu~Lin, and Jinghui Chen. 2023.
\newblock Defending against alignment-breaking attacks via robustly aligned llm.
\newblock \emph{arXiv preprint arXiv:2309.14348}.

\bibitem[{Chao et~al.(2023)Chao, Robey, Dobriban, Hassani, Pappas, and Wong}]{chao2023jailbreaking}
Patrick Chao, Alexander Robey, Edgar Dobriban, Hamed Hassani, George~J Pappas, and Eric Wong. 2023.
\newblock Jailbreaking black box large language models in twenty queries.
\newblock \emph{arXiv preprint arXiv:2310.08419}.

\bibitem[{Dai et~al.(2023)Dai, Pan, Sun, Ji, Xu, Liu, Wang, and Yang}]{dai2023safe}
Josef Dai, Xuehai Pan, Ruiyang Sun, Jiaming Ji, Xinbo Xu, Mickel Liu, Yizhou Wang, and Yaodong Yang. 2023.
\newblock Safe rlhf: Safe reinforcement learning from human feedback.
\newblock \emph{arXiv preprint arXiv:2310.12773}.

\bibitem[{Deng et~al.(2023)Deng, Zhang, Pan, and Bing}]{deng2023multilingual}
Yue Deng, Wenxuan Zhang, Sinno~Jialin Pan, and Lidong Bing. 2023.
\newblock Multilingual jailbreak challenges in large language models.
\newblock \emph{arXiv preprint arXiv:2310.06474}.

\bibitem[{Ding et~al.(2023)Ding, Kuang, Ma, Cao, Xian, Chen, and Huang}]{ding2023wolf}
Peng Ding, Jun Kuang, Dan Ma, Xuezhi Cao, Yunsen Xian, Jiajun Chen, and Shujian Huang. 2023.
\newblock A wolf in sheep's clothing: Generalized nested jailbreak prompts can fool large language models easily.
\newblock \emph{arXiv preprint arXiv:2311.08268}.

\bibitem[{Ganguli et~al.(2022)Ganguli, Lovitt, Kernion, Askell, Bai, Kadavath, Mann, Perez, Schiefer, Ndousse et~al.}]{ganguli2022red}
Deep Ganguli, Liane Lovitt, Jackson Kernion, Amanda Askell, Yuntao Bai, Saurav Kadavath, Ben Mann, Ethan Perez, Nicholas Schiefer, Kamal Ndousse, et~al. 2022.
\newblock Red teaming language models to reduce harms: Methods, scaling behaviors, and lessons learned.
\newblock \emph{arXiv preprint arXiv:2209.07858}.

\bibitem[{Han et~al.(2024)Han, Rao, Ettinger, Jiang, Lin, Lambert, Choi, and Dziri}]{han2024wildguard}
Seungju Han, Kavel Rao, Allyson Ettinger, Liwei Jiang, Bill~Yuchen Lin, Nathan Lambert, Yejin Choi, and Nouha Dziri. 2024.
\newblock Wildguard: Open one-stop moderation tools for safety risks, jailbreaks, and refusals of llms.
\newblock \emph{arXiv preprint arXiv:2406.18495}.

\bibitem[{Huang et~al.(2024)Huang, Wang, Zhang, Xi, An, Wang, and Pan}]{huang2024cross}
Xijie Huang, Xinyuan Wang, Hantao Zhang, Jiawen Xi, Jingkun An, Hao Wang, and Chengwei Pan. 2024.
\newblock Cross-modality jailbreak and mismatched attacks on medical multimodal large language models.
\newblock \emph{arXiv preprint arXiv:2405.20775}.

\bibitem[{Inan et~al.(2023)Inan, Upasani, Chi, Rungta, Iyer, Mao, Tontchev, Hu, Fuller, Testuggine et~al.}]{inan2023llama}
Hakan Inan, Kartikeya Upasani, Jianfeng Chi, Rashi Rungta, Krithika Iyer, Yuning Mao, Michael Tontchev, Qing Hu, Brian Fuller, Davide Testuggine, et~al. 2023.
\newblock Llama guard: Llm-based input-output safeguard for human-ai conversations.
\newblock \emph{arXiv preprint arXiv:2312.06674}.

\bibitem[{Jain et~al.(2023)Jain, Schwarzschild, Wen, Somepalli, Kirchenbauer, Chiang, Goldblum, Saha, Geiping, and Goldstein}]{paraphrase}
Neel Jain, Avi Schwarzschild, Yuxin Wen, Gowthami Somepalli, John Kirchenbauer, Ping-yeh Chiang, Micah Goldblum, Aniruddha Saha, Jonas Geiping, and Tom Goldstein. 2023.
\newblock Baseline defenses for adversarial attacks against aligned language models.
\newblock \emph{arXiv preprint arXiv:2309.00614}.

\bibitem[{Ji et~al.(2024)Ji, Chen, Lou, Hong, Zhang, Pan, Dai, and Yang}]{aligner}
Jiaming Ji, Boyuan Chen, Hantao Lou, Donghai Hong, Borong Zhang, Xuehai Pan, Juntao Dai, and Yaodong Yang. 2024.
\newblock Aligner: Achieving efficient alignment through weak-to-strong correction.
\newblock \emph{arXiv preprint arXiv:2402.02416}.

\bibitem[{Kumar et~al.(2023)Kumar, Agarwal, Srinivas, Li, Feizi, and Lakkaraju}]{erase_and_check}
Aounon Kumar, Chirag Agarwal, Suraj Srinivas, Aaron~Jiaxun Li, Soheil Feizi, and Himabindu Lakkaraju. 2023.
\newblock Certifying llm safety against adversarial prompting.
\newblock \emph{arXiv preprint arXiv:2309.02705}.

\bibitem[{Li et~al.(2024)Li, Dong, Wang, Hu, Zuo, Lin, Qiao, and Shao}]{li2024salad}
Lijun Li, Bowen Dong, Ruohui Wang, Xuhao Hu, Wangmeng Zuo, Dahua Lin, Yu~Qiao, and Jing Shao. 2024.
\newblock Salad-bench: A hierarchical and comprehensive safety benchmark for large language models.
\newblock \emph{arXiv preprint arXiv:2402.05044}.

\bibitem[{Li et~al.(2023{\natexlab{a}})Li, Zhou, Zhu, Yao, Liu, and Han}]{li2023deepinception}
Xuan Li, Zhanke Zhou, Jianing Zhu, Jiangchao Yao, Tongliang Liu, and Bo~Han. 2023{\natexlab{a}}.
\newblock Deepinception: Hypnotize large language model to be jailbreaker.
\newblock \emph{arXiv preprint arXiv:2311.03191}.

\bibitem[{Li et~al.(2023{\natexlab{b}})Li, Wei, Zhao, Zhang, and Zhang}]{li2023rain}
Yuhui Li, Fangyun Wei, Jinjing Zhao, Chao Zhang, and Hongyang Zhang. 2023{\natexlab{b}}.
\newblock Rain: Your language models can align themselves without finetuning.
\newblock \emph{arXiv preprint arXiv:2309.07124}.

\bibitem[{Liu et~al.(2024)Liu, Li, Suh, Vorobeychik, Mao, Jha, McDaniel, Sun, Li, and Xiao}]{liu2024autodan}
Xiaogeng Liu, Peiran Li, Edward Suh, Yevgeniy Vorobeychik, Zhuoqing Mao, Somesh Jha, Patrick McDaniel, Huan Sun, Bo~Li, and Chaowei Xiao. 2024.
\newblock Autodan-turbo: A lifelong agent for strategy self-exploration to jailbreak llms.
\newblock \emph{arXiv preprint arXiv:2410.05295}.

\bibitem[{Liu et~al.(2023)Liu, Xu, Chen, and Xiao}]{liu2023autodan}
Xiaogeng Liu, Nan Xu, Muhao Chen, and Chaowei Xiao. 2023.
\newblock Autodan: Generating stealthy jailbreak prompts on aligned large language models.
\newblock \emph{arXiv preprint arXiv:2310.04451}.

\bibitem[{Markov et~al.(2023)Markov, Zhang, Agarwal, Nekoul, Lee, Adler, Jiang, and Weng}]{markov2023holistic}
Todor Markov, Chong Zhang, Sandhini Agarwal, Florentine~Eloundou Nekoul, Theodore Lee, Steven Adler, Angela Jiang, and Lilian Weng. 2023.
\newblock A holistic approach to undesired content detection in the real world.
\newblock In \emph{Proceedings of the AAAI Conference on Artificial Intelligence}, 12, pages 15009--15018.

\bibitem[{Mazeika et~al.(2024{\natexlab{a}})Mazeika, Phan, Yin, Zou, Wang, Mu, Sakhaee, Li, Basart, Li, Forsyth, and Hendrycks}]{DBLP:conf/icml/MazeikaPYZ0MSLB24}
Mantas Mazeika, Long Phan, Xuwang Yin, Andy Zou, Zifan Wang, Norman Mu, Elham Sakhaee, Nathaniel Li, Steven Basart, Bo~Li, David~A. Forsyth, and Dan Hendrycks. 2024{\natexlab{a}}.
\newblock Harmbench: {A} standardized evaluation framework for automated red teaming and robust refusal.
\newblock In \emph{Forty-first International Conference on Machine Learning, {ICML} 2024, Vienna, Austria, July 21-27, 2024}.

\bibitem[{Mazeika et~al.(2024{\natexlab{b}})Mazeika, Phan, Yin, Zou, Wang, Mu, Sakhaee, Li, Basart, Li et~al.}]{mazeikaharmbench}
Mantas Mazeika, Long Phan, Xuwang Yin, Andy Zou, Zifan Wang, Norman Mu, Elham Sakhaee, Nathaniel Li, Steven Basart, Bo~Li, et~al. 2024{\natexlab{b}}.
\newblock Harmbench: A standardized evaluation framework for automated red teaming and robust refusal.
\newblock In \emph{Forty-first International Conference on Machine Learning}.

\bibitem[{Mehrotra et~al.(2023)Mehrotra, Zampetakis, Kassianik, Nelson, Anderson, Singer, and Karbasi}]{mehrotra2023tree}
Anay Mehrotra, Manolis Zampetakis, Paul Kassianik, Blaine Nelson, Hyrum Anderson, Yaron Singer, and Amin Karbasi. 2023.
\newblock Tree of attacks: Jailbreaking black-box llms automatically.
\newblock \emph{arXiv preprint arXiv:2312.02119}.

\bibitem[{Meta-Llama(2024)}]{promptguard}
Meta-Llama. 2024.
\newblock \href {https://huggingface.co/meta-llama/Prompt-Guard-86M} {Prompt guard 86m}.

\bibitem[{Paulus et~al.(2024)Paulus, Zharmagambetov, Guo, Amos, and Tian}]{paulus2024advprompter}
Anselm Paulus, Arman Zharmagambetov, Chuan Guo, Brandon Amos, and Yuandong Tian. 2024.
\newblock Advprompter: Fast adaptive adversarial prompting for llms.
\newblock \emph{arXiv preprint arXiv:2404.16873}.

\bibitem[{Qi et~al.(2023)Qi, Zeng, Xie, Chen, Jia, Mittal, and Henderson}]{qifine}
Xiangyu Qi, Yi~Zeng, Tinghao Xie, Pin-Yu Chen, Ruoxi Jia, Prateek Mittal, and Peter Henderson. 2023.
\newblock Fine-tuning aligned language models compromises safety, even when users do not intend to!
\newblock In \emph{The Twelfth International Conference on Learning Representations}.

\bibitem[{Robey et~al.(2023)Robey, Wong, Hassani, and Pappas}]{smoothllm}
Alexander Robey, Eric Wong, Hamed Hassani, and George~J Pappas. 2023.
\newblock Smoothllm: Defending large language models against jailbreaking attacks.
\newblock \emph{arXiv preprint arXiv:2310.03684}.

\bibitem[{R{\"o}ttger et~al.(2024)R{\"o}ttger, Kirk, Vidgen, Attanasio, Bianchi, and Hovy}]{rottger2024xstest}
Paul R{\"o}ttger, Hannah Kirk, Bertie Vidgen, Giuseppe Attanasio, Federico Bianchi, and Dirk Hovy. 2024.
\newblock Xstest: A test suite for identifying exaggerated safety behaviours in large language models.
\newblock In \emph{Proceedings of the 2024 Conference of the North American Chapter of the Association for Computational Linguistics: Human Language Technologies (Volume 1: Long Papers)}, pages 5377--5400.

\bibitem[{Wang et~al.(2024)Wang, Huang, Chen, Wang, Pan, Sha, and Huang}]{wang2024blackdan}
Xinyuan Wang, Victor Shea-Jay Huang, Renmiao Chen, Hao Wang, Chengwei Pan, Lei Sha, and Minlie Huang. 2024.
\newblock Blackdan: A black-box multi-objective approach for effective and contextual jailbreaking of large language models.
\newblock \emph{arXiv preprint arXiv:2410.09804}.

\bibitem[{Wei et~al.(2023{\natexlab{a}})Wei, Haghtalab, and Steinhardt}]{DBLP:journals/corr/abs-2307-02483}
Alexander Wei, Nika Haghtalab, and Jacob Steinhardt. 2023{\natexlab{a}}.
\newblock \href {https://doi.org/10.48550/ARXIV.2307.02483} {Jailbroken: How does {LLM} safety training fail?}
\newblock \emph{CoRR}, abs/2307.02483.

\bibitem[{Wei et~al.(2023{\natexlab{b}})Wei, Wang, Li, Mo, and Wang}]{ICD}
Zeming Wei, Yifei Wang, Ang Li, Yichuan Mo, and Yisen Wang. 2023{\natexlab{b}}.
\newblock Jailbreak and guard aligned language models with only few in-context demonstrations.
\newblock \emph{arXiv preprint arXiv:2310.06387}.

\bibitem[{Xie et~al.(2024)Xie, Qi, Zeng, Huang, Sehwag, Huang, He, Wei, Li, Sheng et~al.}]{xie2024sorry}
Tinghao Xie, Xiangyu Qi, Yi~Zeng, Yangsibo Huang, Udari~Madhushani Sehwag, Kaixuan Huang, Luxi He, Boyi Wei, Dacheng Li, Ying Sheng, et~al. 2024.
\newblock Sorry-bench: Systematically evaluating large language model safety refusal behaviors.
\newblock \emph{arXiv preprint arXiv:2406.14598}.

\bibitem[{Xie et~al.(2023)Xie, Yi, Shao, Curl, Lyu, Chen, Xie, and Wu}]{selfreminder}
Yueqi Xie, Jingwei Yi, Jiawei Shao, Justin Curl, Lingjuan Lyu, Qifeng Chen, Xing Xie, and Fangzhao Wu. 2023.
\newblock Defending chatgpt against jailbreak attack via self-reminders.
\newblock \emph{Nature Machine Intelligence}, 5(12):1486--1496.

\bibitem[{Xu et~al.(2024)Xu, Jiang, Niu, Jia, Lin, and Poovendran}]{safedecoding}
Zhangchen Xu, Fengqing Jiang, Luyao Niu, Jinyuan Jia, Bill~Yuchen Lin, and Radha Poovendran. 2024.
\newblock Safedecoding: Defending against jailbreak attacks via safety-aware decoding.
\newblock \emph{arXiv preprint arXiv:2402.08983}.

\bibitem[{Yu et~al.(2023)Yu, Lin, Yu, and Xing}]{yu2023gptfuzzer}
Jiahao Yu, Xingwei Lin, Zheng Yu, and Xinyu Xing. 2023.
\newblock Gptfuzzer: Red teaming large language models with auto-generated jailbreak prompts.
\newblock \emph{arXiv preprint arXiv:2309.10253}.

\bibitem[{Yuan et~al.(2023)Yuan, Jiao, Wang, Huang, He, Shi, and Tu}]{yuan2023toosmart}
Youliang Yuan, Wenxiang Jiao, Wenxuan Wang, Jen-tse Huang, Pinjia He, Shuming Shi, and Zhaopeng Tu. 2023.
\newblock Gpt-4 is too smart to be safe: Stealthy chat with llms via cipher.
\newblock \emph{arXiv preprint arXiv:2308.06463}.

\bibitem[{Zeng et~al.(2024)Zeng, Liu, Mullins, Peran, Fernandez, Harkous, Narasimhan, Proud, Kumar, Radharapu et~al.}]{zeng2024shieldgemma}
Wenjun Zeng, Yuchi Liu, Ryan Mullins, Ludovic Peran, Joe Fernandez, Hamza Harkous, Karthik Narasimhan, Drew Proud, Piyush Kumar, Bhaktipriya Radharapu, et~al. 2024.
\newblock Shieldgemma: Generative ai content moderation based on gemma.
\newblock \emph{arXiv preprint arXiv:2407.21772}.

\bibitem[{Zhang et~al.(2024{\natexlab{a}})Zhang, Cui, Lu, Zhou, Yang, Wang, and Huang}]{zhang2024agent}
Zhexin Zhang, Shiyao Cui, Yida Lu, Jingzhuo Zhou, Junxiao Yang, Hongning Wang, and Minlie Huang. 2024{\natexlab{a}}.
\newblock Agent-safetybench: Evaluating the safety of llm agents.
\newblock \emph{arXiv preprint arXiv:2412.14470}.

\bibitem[{Zhang et~al.(2023{\natexlab{a}})Zhang, Lei, Wu, Sun, Huang, Long, Liu, Lei, Tang, and Huang}]{zhang2023safetybench}
Zhexin Zhang, Leqi Lei, Lindong Wu, Rui Sun, Yongkang Huang, Chong Long, Xiao Liu, Xuanyu Lei, Jie Tang, and Minlie Huang. 2023{\natexlab{a}}.
\newblock Safetybench: Evaluating the safety of large language models with multiple choice questions.
\newblock \emph{arXiv preprint arXiv:2309.07045}.

\bibitem[{Zhang et~al.(2024{\natexlab{b}})Zhang, Lu, Ma, Zhang, Li, Ke, Sun, Sha, Sui, Wang et~al.}]{zhang2024shieldlm}
Zhexin Zhang, Yida Lu, Jingyuan Ma, Di~Zhang, Rui Li, Pei Ke, Hao Sun, Lei Sha, Zhifang Sui, Hongning Wang, et~al. 2024{\natexlab{b}}.
\newblock Shieldlm: Empowering llms as aligned, customizable and explainable safety detectors.
\newblock \emph{arXiv preprint arXiv:2402.16444}.

\bibitem[{Zhang et~al.(2023{\natexlab{b}})Zhang, Wen, and Huang}]{DBLP:conf/acl/ZhangWH23}
Zhexin Zhang, Jiaxin Wen, and Minlie Huang. 2023{\natexlab{b}}.
\newblock \href {https://doi.org/10.18653/V1/2023.ACL-LONG.709} {{ETHICIST:} targeted training data extraction through loss smoothed soft prompting and calibrated confidence estimation}.
\newblock In \emph{Proceedings of the 61st Annual Meeting of the Association for Computational Linguistics (Volume 1: Long Papers), {ACL} 2023, Toronto, Canada, July 9-14, 2023}, pages 12674--12687. Association for Computational Linguistics.

\bibitem[{Zhang et~al.(2024{\natexlab{c}})Zhang, Yang, Ke, Cui, Zheng, Wang, and Huang}]{zhang2024safe}
Zhexin Zhang, Junxiao Yang, Pei Ke, Shiyao Cui, Chujie Zheng, Hongning Wang, and Minlie Huang. 2024{\natexlab{c}}.
\newblock Safe unlearning: A surprisingly effective and generalizable solution to defend against jailbreak attacks.
\newblock \emph{arXiv preprint arXiv:2407.02855}.

\bibitem[{Zhang et~al.(2024{\natexlab{d}})Zhang, Yang, Ke, Cui, Zheng, Wang, and Huang}]{safe_unlearning}
Zhexin Zhang, Junxiao Yang, Pei Ke, Shiyao Cui, Chujie Zheng, Hongning Wang, and Minlie Huang. 2024{\natexlab{d}}.
\newblock Safe unlearning: A surprisingly effective and generalizable solution to defend against jailbreak attacks.
\newblock \emph{arXiv preprint arXiv:2407.02855}.

\bibitem[{Zhang et~al.(2024{\natexlab{e}})Zhang, Yang, Ke, Mi, Wang, and Huang}]{goal_prioritization}
Zhexin Zhang, Junxiao Yang, Pei Ke, Fei Mi, Hongning Wang, and Minlie Huang. 2024{\natexlab{e}}.
\newblock \href {https://arxiv.org/abs/2311.09096} {Defending large language models against jailbreaking attacks through goal prioritization}.
\newblock In \emph{ACL}.

\bibitem[{Zhao et~al.(2024)Zhao, Li, Li, Zhang, and Sun}]{zhao2024defending}
Wei Zhao, Zhe Li, Yige Li, Ye~Zhang, and Jun Sun. 2024.
\newblock Defending large language models against jailbreak attacks via layer-specific editing.
\newblock \emph{arXiv preprint arXiv:2405.18166}.

\bibitem[{Zheng et~al.(2024)Zheng, Yin, Zhou, Meng, Zhou, Chang, Huang, and Peng}]{DRO}
Chujie Zheng, Fan Yin, Hao Zhou, Fandong Meng, Jie Zhou, Kai-Wei Chang, Minlie Huang, and Nanyun Peng. 2024.
\newblock Prompt-driven llm safeguarding via directed representation optimization.
\newblock \emph{arXiv preprint arXiv:2401.18018}.

\bibitem[{Zhou et~al.(2024)Zhou, Wang, Xiong, Xia, Gu, Chai, Zhu, Huang, Dou, Xi et~al.}]{zhou2024easyjailbreak}
Weikang Zhou, Xiao Wang, Limao Xiong, Han Xia, Yingshuang Gu, Mingxu Chai, Fukang Zhu, Caishuang Huang, Shihan Dou, Zhiheng Xi, et~al. 2024.
\newblock Easyjailbreak: A unified framework for jailbreaking large language models.
\newblock \emph{arXiv preprint arXiv:2403.12171}.

\bibitem[{Zou et~al.(2023)Zou, Wang, Kolter, and Fredrikson}]{DBLP:journals/corr/abs-2307-15043}
Andy Zou, Zifan Wang, J.~Zico Kolter, and Matt Fredrikson. 2023.
\newblock \href {https://doi.org/10.48550/ARXIV.2307.15043} {Universal and transferable adversarial attacks on aligned language models}.
\newblock \emph{CoRR}, abs/2307.15043.

\end{thebibliography}
\bibliographystyle{acl_natbib}

\clearpage

\appendix

\section{Implementation Details of Attackers}
\label{app:attacker}

The implementation details of the 13 attackers mentioned in section \ref{appsec:overview_attack} are presented as follows:

\subsubsection{White-box Attacks}

\begin{itemize}
    \item \textbf{GCG \cite{DBLP:journals/corr/abs-2307-15043}}: The Greedy Coordinate Gradient-based Search (GCG) attack method perturbs the input tokens using the loss gradient as an optimization signal. The loss function is designed to maximize the probability of an affirmative prefix, thus guiding the model to produce unsafe or undesirable outputs.
\end{itemize}

\subsubsection{Gray-box Attacks}

\begin{itemize}
    \item \textbf{AutoDAN \cite{liu2023autodan}}: AutoDAN generates jailbreak prompts using a hierarchical genetic algorithm. The goal is to automatically evolve effective attack strategies that bypass the model's safety mechanisms.
    \item \textbf{LAA \cite{andriushchenko2024jailbreaking}}: The LAA method designs adaptive templates and appends adversarial suffixes to the chosen template. The suffix is optimized through random search, achieving high success rates for bypassing model defenses.
    \item \textbf{Advprompter \cite{paulus2024advprompter}}: This method trains an attacker LLM to autoregressively generate adversarial suffixes to a given input query, making it effective for generating successful jailbreak prompts.
\end{itemize}

\subsubsection{Black-box Attacks}

\begin{itemize}
    \item \textbf{GPTFuzzer \cite{yu2023gptfuzzer}}: This method generates new jailbreak templates through iterative mutation of human-written templates. It employs five mutation techniques: generation, crossover, expansion, shortening, and rephrasing, all aimed at finding prompts that can successfully bypass model defenses.
    \item \textbf{Cipher \cite{yuan2023toosmart}}: The Cipher attack works by encoding instructions in a cryptic manner, such that the model's safety alignment mechanisms fail to interpret the instructions correctly, enabling a jailbreak.
    \item \textbf{DeepInception \cite{li2023deepinception}}: This attack creates diverse scenes and characters to mislead the target model’s safety filters, thus circumventing its safety alignment.
    \item \textbf{In-context Learning Attacks \cite{ICD}}: In-context learning exploits few-shot demonstrations to manipulate the target model's behavior and trigger unsafe outputs.
    \item \textbf{Jailbroken \cite{DBLP:journals/corr/abs-2307-02483}}: The Jailbroken attack method targets two key failure modes in LLM safety alignment: competing objectives and mismatched generalization. By leveraging these weaknesses, it crafts prompts that can bypass the model’s safety mechanisms.
    \item \textbf{MultiLingual \cite{deng2023multilingual}}: This attack translates input queries into low-resource languages, often evading the safety mechanisms in models that are less robust in non-major languages, thus achieving a successful jailbreak.
    \item \textbf{PAIR \cite{chao2023jailbreaking}}: PAIR involves using an attacker LLM to iteratively refine jailbreak prompts, enhancing their effectiveness through repeated refinement.
    \item \textbf{ReneLLM \cite{ding2023wolf}}: ReneLLM combines two techniques: prompt rewriting and scenario nesting. These methods are used to reframe the input in ways that bypass the model’s defenses.
    \item \textbf{TAP \cite{mehrotra2023tree}}: The Tree of Attacks with Pruning (TAP) method maintains a structured flow in the form of a tree to iteratively optimize the jailbreak prompt. The attack continues until a successful jailbreak prompt is found.
\end{itemize}

\section{Implementation Details of Defenders}
\label{app:defender}

\begin{figure*}[!t]
    \centering
    \includegraphics[width=\linewidth]{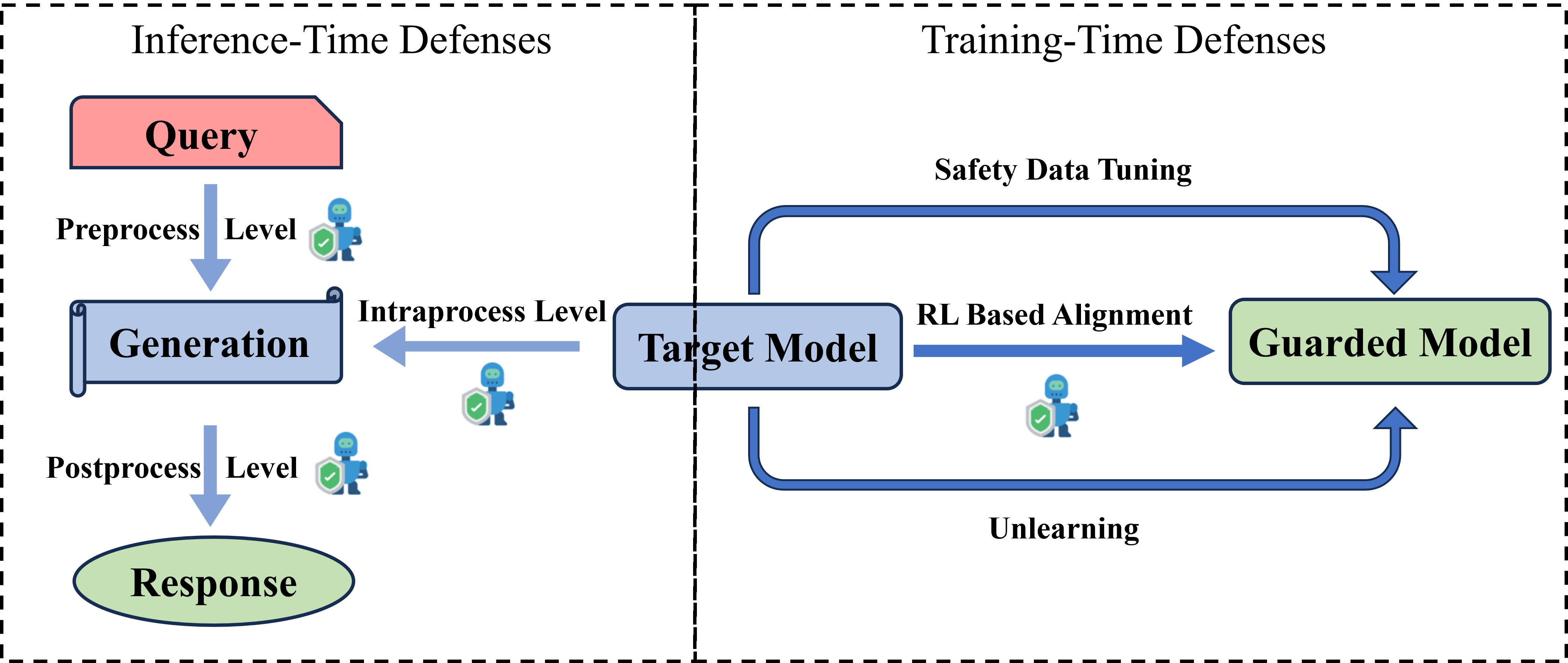}
    \caption{Overview of the categorization of safety defense methods.}
    \label{fig:defense_overview}
\end{figure*}
\begin{table*}[!t]
\centering
\small
\renewcommand{\arraystretch}{1.2}
\setlength{\tabcolsep}{8pt} 
\resizebox{\linewidth}{!}{
\begin{tabularx}{\textwidth}{|m{1.5cm}|m{4cm}|m{3.0cm}|m{5.28cm}|} 
\hline
\textbf{Type} & \textbf{Description} & \textbf{Usage} & \textbf{Methods} \\
\hline
\textbf{Preprocess} & 
The input is preprocessed to detect or guard against harmful content before the generation process. & 
defend(self, message) $\rightarrow$ (defended\_message, if\_reject) & 
\makecell[l]{PPL \cite{PPL} \\ Self Reminder \cite{selfreminder} \\ Prompt Guard \cite{promptguard} \\ Goal Prioritization \cite{goal_prioritization}\\ Paraphrase \cite{paraphrase}\\ ICD \cite{ICD}} \\
\hline
\textbf{Intraprocess} & 
Utilizes safer decoding or generation strategies to ensure the outputs are secure and adhere to safety guidelines. & 
defend(self, model, messages) $\rightarrow$ response & 
\makecell[l]{SmoothLLM \cite{smoothllm}\\ SafeDecoding \cite{safedecoding} \\ DRO \cite{DRO} \\ Erase and Check \cite{erase_and_check} \\ Robust Aligned \cite{cao2023defending} } \\
\hline
\textbf{Postprocess} & 
The output is postprocessed to enhance safety. & 
defend(self, response) $\rightarrow$ defended\_response & 
\makecell[l]{Self Evaluation \cite{self-evaluation} \\ Aligner \cite{aligner}} \\
\hline
\end{tabularx}
}
\caption{Overview of Inference-Time Defense Types and Their Components}
\label{tab:defenders}
\end{table*}

\subsection{Inference Time Defense}
\label{app:defender1}

We selected 13 inference-time defense methods, categorized into three main strategies, and provide a unified interface for these defense methods, which can be utilized in the \textit{defend\_chat} function, as illustrated in Table \ref{tab:defenders}.

\subsection{Training Time Defense}
\label{app:defender2}
The Training Time Defense is applied during the training process of large language models. It includes three categories: Safety Data Tuning, RL-based Alignment and Unlearning. And we implement a method for each category.

\begin{itemize}
     \item \textbf{Safety Data Tuning}: Safety-Tuning \cite{safe_tuning}
     \item \textbf{RL-based Alignment}: Safe RLHF \cite{dai2023safe}
     \item \textbf{Unlearning}: Safe Unlearning \cite{safe_unlearning}
\end{itemize}

Organized into key directories such as config, data, model, train, and utils, the training framework ensures modularity and ease of management. The data directory handles preprocessing and dataset management, while train encompasses specialized subfolders for different training methods, including trainers and workflows, which is flexible and customizable.

\section{Implementation Details of Scorers}
\label{app:scorer}

The implementation details of the seven scorers mentioned in section \ref{appsec:evaluation} are presented as follows:

\begin{itemize}
    \item \textbf{PatternScorer.} This scorer judges the success of a jailbreak attempt by examining whether the response matches predefined jailbreaking failure patterns. A response is considered safe if it includes a certain failure pattern.
    \item \textbf{PrefixMatchScorer.} This scorer judges the success of a jailbreak by checking the prefix of the response. A response is considered unsafe if it starts with a jailbreaking target.
    \item \textbf{ClassficationScorer}. This scorer uses a classification model to assess the safety of the response. The scorer is initialized by a Roberta model from \citet{yu2023gptfuzzer}, and it is also acceptable to init this scorer with any customized classification model.
    \item \textbf{ShieldLMScorer.} This scorer uses \texttt{ShieldLM} to assess the safety of the response, which incorporates an analysis alongside its judgment explaining the reason behind its decision.
    \item \textbf{HarmBenchScorer.} This scorer uses the \texttt{HarmBench-Llama-2-13b-cls} classification model \cite{mazeikaharmbench} to assess the safety of the response.
    \item \textbf{LlamaGuard3Scorer.} This scorer uses \texttt{Llama-Guard-3-8B} to assess the safety of the response. It also provides the unsafe category if the response is judged as unsafe.
    \item \textbf{PromptedLLMScorer.} This scorer prompts a model to assess the safety of the response. We incorporate four judge prompts from \citet{qifine}, \citet{zhang2024shieldlm}, \citet{mehrotra2023tree} and \citet{chao2023jailbreaking}.
\end{itemize}

\end{document}